\def\year{2019}\relax
\documentclass[letterpaper]{article} 
\usepackage{aaai19}  
\usepackage{times}  
\usepackage{helvet}  
\usepackage{courier}  
\usepackage{url}  
\usepackage{graphicx}  

\usepackage[spacesep,definevectors]{easyvector}
\usepackage{amsmath}
\usepackage{mathtools}
\usepackage{xspace}
\usepackage{multirow}

\DeclareMathOperator*{\argmax}{argmax}
 
\newcommand{\inprod}[2]{\langle #1, #2\rangle} 
\newcommand{\Relu}{\mathsf{ReLU}} 

\newcommand{\Direct}{\textsf{Direct}}
\newcommand{\Distiller}{\textsf{Distiller}}
\newcommand{\spcA}{\textsf{SYN}\xspace{}}
\newcommand{\spcB}{\textsf{ANT}\xspace{}}

\begin{document}


\title{Antonym-Synonym Classification Based on New Sub-space Embeddings}

\author{
    Muhammad Asif Ali, \textsuperscript{1} 
    Yifang Sun, \textsuperscript{1}
    Xiaoling Zhou, \textsuperscript{1}
    Wei Wang, \textsuperscript{1, 2}
    Xiang Zhao \textsuperscript{3}\\
    \textsuperscript{1} School of Computer Science and Engineering, UNSW, Australia\\
    \textsuperscript{2} College of Computer Science and Technology, DGUT, China\\
    \textsuperscript{3} Key Laboratory of Science and Technology on Information System Engineering, NUDT, China\\
    muhammadasif.ali@unsw.edu.au, \{yifangs, xiaolingz, weiw\}@cse.unsw.edu.au, xiangzhao@nudt.edu.cn\\
}

\maketitle


\begin{abstract}
    Distinguishing antonyms from synonyms is a key challenge for many NLP applications focused on the lexical-semantic relation extraction. Existing solutions relying on large-scale corpora yield low performance because of huge contextual overlap of antonym and synonym pairs. We propose a novel approach entirely based on pre-trained embeddings. We hypothesize that the pre-trained embeddings comprehend a blend of lexical-semantic information and we may distill the task-specific information using Distiller, a model proposed in this paper. Later, a classifier is trained based on features constructed from the distilled sub-spaces along with some word level features to distinguish antonyms from synonyms. Experimental results show that the proposed model outperforms existing research on antonym synonym distinction in both speed and performance.
    
\end{abstract}


\section{Introduction}

Distinguishing between antonymy and synonymy is one of the crucial problems for NLP applications, especially those focused on lexical-semantic relation extraction, such as sentiment analysis, semantic relatedness, opinion mining and machine translation. 
We define synonyms as semantically similar words (e.g., \texttt{disperse} and \texttt{scatter}) and antonyms as highly contrasting words (e.g., \texttt{disperse} and \texttt{garner})~\cite{DBLP:conf/naacl/OnoMS15}. Existing manually curated lexical resources (e.g.,
WordNet~\cite{fellbaum1998wordnet}) are unable to address this problem, due to
the limited coverage. This calls the need for a machine learning model to classify a
given pair of words as either synonyms or antonyms.

Traditional research on antonym/synonym distinction makes use of word embeddings to capture the semantic relatedness among antonym/synonym pairs based on co-occurrence statistics~\cite{DBLP:conf/ijcnlp/ScheibleWS13,DBLP:conf/naacl/OnoMS15,DBLP:conf/acl/NguyenWV16,DBLP:conf/rep4nlp/Vulic18}. However, the embedding models have a tendency to mix different lexico-semantic relations, so the performance cannot be guaranteed when they are applied on specific lexical-semantic analysis tasks~\cite{DBLP:conf/naacl/GlavasV18}. This situation worsens in the case of antonym and synonym distinction, because these words can be used interchangeably and are considered indistinguishable~\cite{DBLP:journals/coling/MohammadDHT13}. This is verified in this paper by the poor performance of a baseline classifier, i.e., \Direct{}, that is purely designed based on pre-trained word embeddings.

Recently, pattern based approaches have gained considerable research attention for lexical-semantic relation extraction~\cite{DBLP:conf/acl/RothW14,DBLP:conf/conll/SchwartzRR15}. 
\citeauthor{DBLP:conf/eacl/NguyenWV17} \shortcite{DBLP:conf/eacl/NguyenWV17} formulated special
lexico-syntactic pattern features that concatenate the lemma, POS-tag and dependency label along the dependency path. Their model achieved the state-of-the-art result on antonym/synonym distinction task. Pattern based methods yield a low recall owing to the lexical variations and the sparsity of lexico-syntactic patterns, which limits the information sharing among semantically similar patterns. Existing attempts to resolve the sparsity via generalization cause the resultant patterns to be highly overlapping across different classes, which has a detrimental effect on the classification accuracy. Moreover, they have to use a large-scale text corpus to combat sparsity, which drastically increases the computational overhead.

In this paper, we propose a novel two-phase approach to address the above-mentioned challenges for antonym/synonym distinction by eliminating the need for a large text corpus. Firstly, we use a new model named: \Distiller{}, which is a set of non-linear encoders, to distill task-specific information from \emph{pre-trained word embeddings} to dense sub-spaces. For this, we design two new loss functions to capture unique relation-specific properties for antonyms and synonyms, i.e., symmetry, transitivity and the special \emph{trans-transitivity} (explained in section~\ref{formal_def}), in two different sub-spaces namely: \spcA{} for synonyms and \spcB{} for antonyms. Finally, a classifier is trained using features constructed from the distilled information, in addition to other word-level features, to distinguish antonym and synonym pairs.

Note that our problem formulation is same as that of existing works 
with the distinction that we replace the requirement of a large-scale 
corpus by the availability of pre-trained word embeddings. This makes 
our setting more appealing and flexible, as pre-trained embeddings 
are widely available (e.g., word2vec, Glove, and ELMo), and they have 
arguably high coverage and quality due to the 
gigantic training corpus. In addition, they are available 
for many languages \cite{DBLP:journals/corr/JoulinGBDJM16}, and are easily adaptable 
as one can customize the pre-trained embeddings by further training 
with domain-specific corpus~\cite{DBLP:conf/acl/RotheS16,DBLP:journals/corr/abs-1710-10280,DBLP:conf/naacl/VulicM18}. We summarize the major contributions of this paper as follows: 

\begin{itemize}
\item We propose a novel model for the antonym/synonym distinction. Compared with existing research, our model makes less stringent data requirements (only requiring pre-trained embeddings), hence it is more practical and efficient. 
  \item We introduce \Distiller{}: a set of non-linear encoders to distill task-specific information from pre-trained embeddings in a performance-enhanced fashion. In addition, we propose new loss functions to enforce the distilled representations to capture relation-specific properties.
  
\item We demonstrate the effectiveness of the proposed model by comprehensive experimentation. Our model outperforms the existing research on antonym/synonym distinction by a large margin.

\item We construct a new dataset for antonym/synonym distinction task in Urdu language to demonstrate the language-agnostic properties of the proposed model.  
\end{itemize}


\section{Related Work}
\label{related_work}

Existing research on antonym/synonym distinction can be classified as (i) embeddings based and (ii) pattern based approaches. 

\subsubsection{Embeddings Based Approaches}

The embeddings based approaches rely on the distributional hypothesis, i.e.,
\emph{the words with similar (or opposite) meanings appear in a similar context}~\cite{DBLP:journals/jair/TurneyP10}. These models are trained using neural networks~\cite{DBLP:journals/corr/abs-1301-3781,DBLP:conf/nips/MikolovSCCD13} or matrix factorization~\cite{DBLP:conf/emnlp/PenningtonSM14}. 
Dominant embeddings based approaches rely on training embedding vectors 
using different features extracted from large scale text corpora. For example, 
\citeauthor{DBLP:conf/ijcnlp/ScheibleWS13}~\shortcite{DBLP:conf/ijcnlp/ScheibleWS13} 
explained the distributional differences between antonyms and synonyms.  
\citeauthor{DBLP:conf/emnlp/AdelS14}~\shortcite{DBLP:conf/emnlp/AdelS14} employed
co-reference chains to train skip-gram model to distinguish antonyms. 
\citeauthor{DBLP:conf/acl/NguyenWV16}~\shortcite{DBLP:conf/acl/NguyenWV16}
integrated distributional lexical contrast information in the skip-gram model for antonym and synonym distinction. 

The supervised variants of the embeddings based models employ existing resources, i.e., thesaurus in combination with the distributional information for the distinction task. \citeauthor{DBLP:conf/acl/PhamLB15}~\shortcite{DBLP:conf/acl/PhamLB15} introduced multi-task lexical contrast by augmenting the skip-gram model with supervised information from WordNet.
\citeauthor{DBLP:conf/naacl/OnoMS15}~\shortcite{DBLP:conf/naacl/OnoMS15} proposed a model that uses distributional information alongside thesaurus to detect probable antonyms. The major limitation of the embeddings based methods is their inability to discriminate between different lexico-semantic relations; e.g., in Glove the top similar 
words for the word \texttt{small} yield a combination of synonyms (e.g.,
\texttt{tiny}, \texttt{little}), antonyms (e.g., \texttt{large}, \texttt{big}), and irrelevant terms (e.g., \texttt{very}, \texttt{some}).

\subsubsection{Pattern Based Approaches}

The pattern based approaches rely on capturing lexico-syntactic patterns from large scale text corpora. For example, \citeauthor{DBLP:conf/ijcai/LinZQZ03}~\shortcite{DBLP:conf/ijcai/LinZQZ03} 
proposed two patterns, i.e., \emph{either X or Y} and \emph{from X to Y} particularly
indicative of antonym pairs. \citeauthor{DBLP:conf/acl/RothW14}~\shortcite{DBLP:conf/acl/RothW14} 
used discourse markers in combination with the lexico-syntactic patterns to distinguish antonyms 
from synonyms. \citeauthor{DBLP:conf/conll/SchwartzRR15}~\shortcite{DBLP:conf/conll/SchwartzRR15} used
symmetric patterns to assign different vector representations to the synonym 
and antonym pairs. \citeauthor{DBLP:conf/eacl/NguyenWV17}~\shortcite{DBLP:conf/eacl/NguyenWV17} integrated the lexico-syntactic pattern features with the distributional information.

A major limitation of the pattern based methods is the huge overlap in feature space (i.e., lexico-syntactic patterns) across different classes, which hinders improvement in performance. For example, a commonly confused pattern is the noun-compound, which can be used to represent both the synonym pairs (\texttt{government administration}; \texttt{card board}) and the
antonym pairs (\texttt{graduate student}; \texttt{client server}).


\section{The Proposed Model}
\label{Proposed_model}

\subsection{Problem Definition}

In this paper, we aim to build a model that can classify a given pair of words as either 
synonyms or antonyms. We consider synonyms as semantically similar words (having 
similar meanings), and antonyms as highly contrasting words (having opposite meanings).
Similar to the previous
works~\cite{DBLP:conf/acl/RothW14,DBLP:conf/conll/SchwartzRR15,DBLP:conf/eacl/NguyenWV17},
we assume the availability of training dataset, i.e., pairs of words with class labels 
indicating the pair satisfies either synonym or antonym relation. In contrast to the 
existing research, the proposed model no longer relies on a text corpus. We replace 
it with the available pre-trained word embeddings.

\subsection{Overview}
Our proposed model consists of two phases: in \emph{Phase I}, we train the 
\Distiller{} to distill task-specific representations of all words in the 
pre-trained embeddings via non-linear projections; in \emph{Phase II}, we 
train a classifier that exploits features constructed from the 
distilled representations together with other word-level features to classify a given 
pair of words into either synonyms or antonyms. Both phases are trained in 
a supervised fashion using the same set of training instances.

We argue that the proposed two-phase design has the following advantages: (i) it allows us to answer the question, \textit{whether we can collect enough information from pre-trained embeddings for synonym/antonym classification}, and (ii) it provides us with the maximal flexibility to accommodate many other types of features, including corpus-level features, in our future work.

\subsection{Phase I (\Distiller{})}
\label{formal_def}
\Distiller{} uses two different neural-network encoders to project pre-trained 
embeddings to two new sub-spaces in a non-linear fashion. Each encoder is a 
feed-forward neural network with two hidden layers using sigmoid as the activation 
function. Mathematically, we model them as encoder functions; 
we use $enc_S(v)$ and $enc_A(v)$ for projections of word $v$'s pre-trained 
embedding vector to the \emph{\underline{s}ynonym and \underline{a}ntonym sub-spaces},
respectively. 

We observe that antonyms and synonyms are special kind of relations (denoted as $r_A$ and $r_S$, respectively), that exhibit some unique properties. Synonyms possess \emph{symmetry} and \emph{transitivity}, whereas, antonyms possess \emph{symmetry} and \emph{trans-transitivity}.

\begin{figure}[ht]
    \centering
    \includegraphics[scale = 0.15]{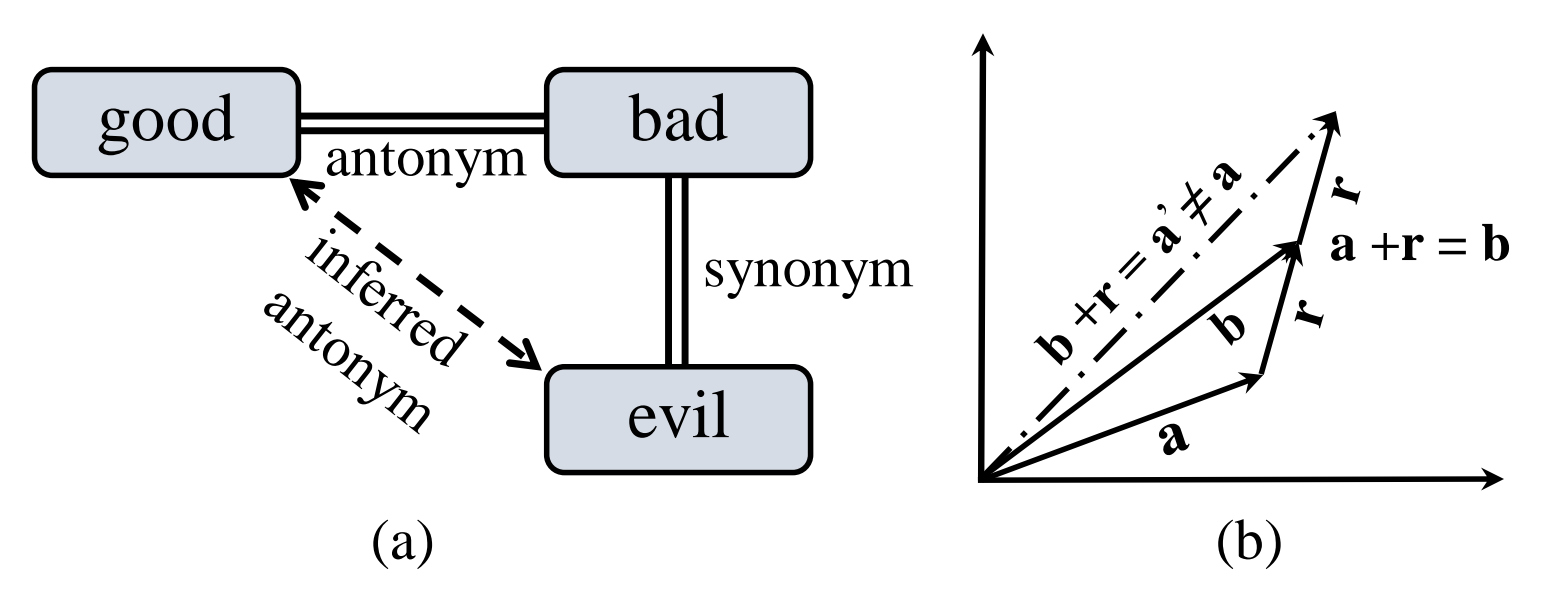} 
    \caption{(a) Illustrating the trans-transitivity of antonym and synonym relations, (b) Limitation of the translational embeddings to capture symmetric relations}
    \label{fig:transtive}
\end{figure}

\emph{Symmetry} implies that $r(a, b)$ \emph{if and only if} $r(b, a)$, where $r(a, b)$ means $a$ and $b$ participate in the relation $r$.
For synonyms, the \emph{transitivity} implies that \emph{if $r_{S}(a,b)$ and 
$r_{S}(b,c)$ then $r_{S}(a,c)$ also holds}. 
For antonyms, the \emph{trans-transitivity} implies that \emph{if $r_{A}(a,b)$ and $r_{S}(b,c)$ then $r_{A}(a,c)$} --- as
shown by the inferred antonym relation between \emph{good} and \emph{evil}
in Figure~\ref{fig:transtive}(a). Trans-transitivity is a unique property that helps to infer probable antonym pairs.

We note that the existing translational entity embedding 
models~\cite{DBLP:conf/nips/BordesUGWY13,DBLP:conf/naacl/YoonSPP16} are not an appropriate choice for this problem. In translational embedding models, a relation $r$ is modeled as a translation vector $\rr$, and $r(a, b)$ implies the $L_1$ or $L_2$-norm of $({\aa + \rr - \bb})$ is small. As shown in Figure~\ref{fig:transtive}(b), it is not possible for the translational embeddings to preserve symmetry by accommodating both vector operations $\aa + \rr = \bb$ and $\bb + \rr = \aa$ at the same time. 

Formally, if we try to model a symmetric relation (i.e., synonyms or antonyms) via the translational model, we have
\begin{align*}
\left.
\begin{aligned}
\aa + \rr - \bb = \Bepsilon_1 \quad\\
\bb + \rr - \aa = \Bepsilon_2 \quad
\end{aligned}
\right\}
\Rightarrow \quad
2\rr = \Bepsilon_1 + \Bepsilon_2 
\end{align*}
It means $\rr$ must be small to be able to model both $r(a, b)$ and $r(b, a)$.
This leads to huge challenge in distinguishing different $\rr$s, or has the risk of modeling one of the two pairs badly. A similar explanation suffices to expose the limitation of translational embeddings to preserve transitivity.

While the translational embeddings fail to capture the antonym and synonym relation pairs, we propose a model that:
\begin{itemize}
    \item for each word vector, creates two distilled embeddings in two sub-spaces (i.e., \spcA{} and \spcB{}).
    \item models the symmetry and transitivity preserving synonym relation in \spcA{}.
    \item models the symmetry and trans-transitivity preserving antonym relation using both \spcB{} and \spcA{}.
\end{itemize}

\subsubsection{Loss Function for Synonyms}
\label{sec:loss-funct-synonyms}
In order to capture synonym relations in \spcA{}, we use a margin based loss to embed the vectors corresponding to synonym pairs (positive examples) 
close to each other, while at the same time pushing apart the vectors for irrelevant 
pairs (negative examples). This formulation preserves the symmetry and the transitivity of synonym pairs in the \spcA{} sub-space. While the symmetry is enforced by the commutative nature of the inner product of real
vectors\footnote{Inner product can be regarded as some notion of similarity.},\label{key}
the transitivity is preserved by following justification: if $\aa$ is
embedded close to $\bb$ and $\bb$ is embedded close to $\cc$, then $\aa$ has a
high chance of being embedded close to $\cc$. 
It, moreover, ensures that we can distinguish between the synonym pairs and the 
irrelevant pairs within the \spcA{} sub-space.

The loss function for modeling the synonyms is shown in Equation~\eqref{eq:SYN}. 
Note that only the embeddings corresponding to the \spcA{} sub-space are involved. %

\begin{equation}
  \label{eq:SYN}
\begin{aligned}
  L_S &= \sum_{(a, b) \in T_S}  \Relu(1 - f(a,b)) \\
    &+ \sum_{(a', b') \in T'_S} \Relu(1 + f(a',b'))
\end{aligned} 
\end{equation}
where $f(a, b) = \tanh(\inprod{enc_S(a)}{enc_S(b)})$, $T_S$ is the set of synonym training instances, $T'_S$ is the set of negative pairs, and $enc_S(v)$ is the non-linear mapping for the words in the vocabulary $V$ to distilled embedding in the $\spcA$ sub-space. 

$T'_S$ is generated from $T_S$ by repeating the following corruption process 
$k=5$ times: randomly pick $a$ or $b$ from $T_S$, and replace it with another 
randomly sampled word from training vocabulary. %

\subsubsection{Loss Function for Antonyms}
\label{sec:loss-funct-antonyms}

In order to capture the symmetry and trans-transitivity for antonym pairs,
our idea is to \emph{correlate} the \spcB{} and \spcA{} sub-spaces. 

For illustration, consider an antonym pair $r_{A}(a,b)$, and let $b_i$ be 
any synonym of $b$. When the fact that $b_i$ and $b$ are synonyms is 
sufficiently learned in \spcA{}, we expect $\bb_i$ to be close to $\bb$, later,
we encourage the embedding vector $\aa \in \spcB$ to be close to $\bb \in \spcA$. As a result, it
is highly likely that $\aa \in \spcB$ will also be close to $\bb_i \in \spcA$.

In order to capture the trans-transitivity using the \Distiller{}, for a given 
antonym relation pair $r_{A}(a,b)$, we bring the vector $\aa \in \spcB$ close to the vector $\bb \in \spcA$.
For symmetry, we augment the training data by swapping the relation pairs $r_{A}(a,b)$ to get $r_{A}(b,a)$, and correspondingly update the embeddings for the vectors $\bb \in \spcB$ and $\aa \in \spcA$. This design preserves the symmetry and the trans-transitivity property for the antonym pairs using both \spcA{} and \spcB{} sub-spaces. The loss function for modeling antonyms is illustrated in Equation~\eqref{eq:ANTEQ}. Note that unlike Equation~\eqref{eq:SYN}, for this loss function, the embeddings for both the sub-spaces are involved.

\begin{equation}
  \label{eq:ANTEQ}
\begin{aligned} 
 L_A &= \sum_{(a, b) \in T_A}  \Relu(1 - g(a,b)) \\
& + \sum_{(a', b') \in T'_A} \Relu(1 + g(a',b'))
\end{aligned}
\end{equation}
where $g(a, b) = \tanh(\inprod{enc_A(a)}{enc_S(b)})$, $T_A$ is the set of the antonym 
training instances, $T'_A$ is the set of negative pairs, $enc_S(v)$ is defined as 
in Equation~\eqref{eq:SYN} and $enc_A(v)$ is the non-linear mapping for the words in the vocabulary $V$ to distilled embedding in the $\spcB$ sub-space.

$T'_A$ is generated from $T_A$ by repeating the following corruption process $k=5$ times: randomly pick $a$ or $b$ from $T_A$, and replace it with another randomly sampled word from the training vocabulary.

\subsubsection{Synonymy and Antonymy Scores}
For a given word pair $(a,b)$, we use our distilled embeddings to compute its \emph{synonymy 
score} as $cos(enc_{S}(a),enc_S(b))$ and the \emph{antonymy score} as 
$\max(\cos(enc_A(a), enc_S(b)),$ $ \cos(enc_A(b), enc_S(a)))$.

\subsubsection{Loss Function for \Distiller}
\label{sec:loss-funct-dist}
The loss function of the \Distiller{} is $L_S + L_A + L_M $, where $ L_S$ and $L_A$ have been defined in Equations~\eqref{eq:SYN} and \eqref{eq:ANTEQ} respectively. $L_M$ is the cross-entropy loss defined in Equation~\eqref{CE_eq}.

\begin{equation}
\label{CE_eq}
L_M = -\frac{1}{N} \sum_{i=1}^{N} \log({\hat{p}}(y_i|a_i, b_i))
\end{equation}

For $L_M$, we concatenate the synonymy score $\xx_1$ and the antonymy score $\xx_2$ from the distilled embeddings to form the feature vector $\xx = [\xx_1, \xx_2]$, and use softmax function (Equation~\eqref{softmax_eq}) to predict the class label.

\begin{equation}
\label{softmax_eq}
\begin{aligned}
{\hat{p}}(\yy \mid (\aa,\bb)) & = softmax(\WW\xx+\hat{\bb})\\
{\hat{y}} & = \argmax_y {\hat{p}}(\yy|\aa,\bb)
\end{aligned}
\end{equation}

where $\WW$ is the weight matrix and $\hat{\bb}$ is the bias. We optimize the loss of the \Distiller{}, in an end-to-end fashion.

\subsection{Phase II (Classifier)}

In Phase II, we train a classifier to distinguish the synonym and antonym pairs. We use the XGBoost classifier~\cite{DBLP:conf/kdd/ChenG16}, and employ following
features.

\subsubsection{New Features from \Distiller{}}%
For a given test pair $(a,b)$, we compute two new features exploiting our
distilled embeddings (1) \emph{synonymy score} and (2) 
\emph{antonymy score}.

\subsubsection{Features from Pre-trained Embeddings}%
Similar to the best performing solutions for antonym and synonym distinction that 
consider a combination of lexico-syntactic patterns and the distributional embeddings, we use distributional similarity score, i.e., $cos(\aa,\bb)$ as a classification feature, where $\aa$ and $\bb$ correspond to the \emph{pre-trained} word embedding vectors. It helps in capturing highly confident antonym and/or synonym pairs trained using distributional information. In Table~\ref{tab:Feature_Analysis}, 
this feature is labeled as: \emph{distributional}.

\subsubsection{Features from Negation Prefix}%
Words that differ only by one of the known negation prefixes are highly unlikely 
to be synonym pairs, e.g., \texttt{(able,unable), (relevant,irrelevant)} etc. 
Similar to ~\citeauthor{DBLP:conf/starsem/RajanaCAS17} \shortcite{DBLP:conf/starsem/RajanaCAS17}, we use the following set of negation prefixes:
\textit{\{de, a, un, non, in, ir, anti, il, dis, counter, im, an, sub, ab\}}. 
We use a binary feature for the candidate pairs, where the feature value is 1 
only if the two words in the candidate pair differ by one of the negation prefix. 
In Table~\ref{tab:Feature_Analysis}, this feature is labeled as: \emph{prefix}.


\section{Experiments}
\label{experimentions}
\subsection{Dataset}

For model training, we use an existing dataset previously used by~\cite{DBLP:conf/conll/SchwartzRR15,DBLP:conf/acl/RothW14,DBLP:conf/acl/NguyenWV16,DBLP:conf/eacl/NguyenWV17}.
It has been accumulated from different sources encompassing WordNet~\cite{fellbaum1998wordnet} and
WordNik\footnote{\url{https://www.wordnik.com/}}. The details of the dataset are given in Table~\ref{tab:dataset_ANTSYN}, it contains antonym and synonym pairs for three categories (i.e., verbs, adjectives and nouns) in the ratio (1:1).
In order to come up with a unanimous platform for comparative evaluation, we use the priorly 
defined data splits by the existing models to training, test and dev sets. 
The training data is used to train the \Distiller{} in Phase-I and the classifier in Phase-II. The development data is used for \Distiller{}'s parameter
tuning. The model performance is reported on the test set.

\begin{table}[htbp]
    \centering
    \resizebox{0.8\linewidth}{!}{
    \begin{tabular}{l | rrrr}
        \textbf{Category} & \textbf{Train} & \textbf{Dev} & \textbf{Test} & \textbf{Total}\\
        \hline
        Verb & 2534 & 182 & 908 & 3624\\
        Noun & 2836 & 206 & 1020 & 4062 \\
        Adjective & 5562 & 398 & 1986 & 7946\\
        \hline
    \end{tabular}}
    \caption{Antonym/Synonym distinction dataset}
    \label{tab:dataset_ANTSYN}
\end{table}

\subsection{Experimental Settings}
\label{experimental_settings}

One advantage of our model is its ability to work with any set of word embeddings. For experimentation, we use following set of pre-trained embeddings:
\begin{enumerate}
\item \textbf{Random Vectors} We use 300d random vectors.
\item \textbf{Glove}~\cite{DBLP:conf/emnlp/PenningtonSM14}, purely unsupervised word embeddings. We use 300d pre-trained Glove embeddings.
\item \textbf{dLCE}~\cite{DBLP:conf/acl/NguyenWV16}, customized embeddings for antonym synonym distinction task. Its dimensionality is 100d.
\item \textbf{ELMO}~\cite{DBLP:conf/naacl/PetersNIGCLZ18}, deep contextualized embeddings based on character sequences. Its dimensionality is 1024d.
\end{enumerate}

Note that our model does not rely on a text corpus, so for the ELMO, 
we only consider the character sequence of the words in our dataset to acquire the embedding vectors. For Glove and dLCE, the vectors corresponding to the out-of-vocabulary tokens were randomly initialized. We use a smaller and compact representation for the sub-spaces, as they are supposed to preserve only the task-specific information. The dimensionality of each sub-space, (i.e., \spcB{} and \spcA{} ) is set to 60d. The neural network encoders used in the \Distiller{} employ 80 units in the first layer and 60 units in the second layer. We use the Adam-Optimizer~\cite{DBLP:journals/corr/KingmaB14} to train the \Distiller{}. All the experiments are performed on Intel Xenon Xeon(R) CPU E5-2640 (v4) with 256 GB main memory and Nvidia 1080Ti GPU.

\subsection{Baseline Models / Model Comparison}
\label{baselines__}

We compare our work against the following models.

\subsubsection{\Direct{} Baseline.}
The \Direct{} baseline is used to show the performance of using only the pre-trained embeddings to distinguish the antonym and synonym pairs. For \Direct{} baseline, we cluster the vector difference for antonym and synonym pairs using k-means clustering to get k-pivot vectors as 
representatives of synonym and/or antonym candidate pairs. For classification, 
we use an XGBoost classifier with (i) cosine similarity, (ii) distance to the pivot 
vectors and (iii) vector difference as features.

\subsubsection{Discourse Markers.} \citeauthor{DBLP:conf/acl/RothW14} \shortcite{DBLP:conf/acl/RothW14} used discourse markers (indicators of relations) alongside lexico-syntactic patterns to design vector space models. For the given data set, Michel Roth has already computed the performance of his methods~\cite{DBLP:conf/eacl/NguyenWV17}, we use the same scores for comparison. 

\subsubsection{Symmetric Patterns.} \citeauthor{DBLP:conf/conll/SchwartzRR15} \shortcite{DBLP:conf/conll/SchwartzRR15} 
used automated techniques to extract symmetric patterns (sequence of 3-5 tokens encompassing two wild-cards and 1-3 tokens) from plain text. We use 500d embeddings (from author's web-page \footnote{\url{https://homes.cs.washington.edu/~roysch/papers/sp_embeddings/sp_embeddings.html}}). Similar to \cite{DBLP:conf/eacl/NguyenWV17}, we calculate cosine similarity between the test pair and use an SVM classifier to categorize them.

\subsubsection{AntSynNET.} \citeauthor{DBLP:conf/eacl/NguyenWV17} 
\shortcite{DBLP:conf/eacl/NguyenWV17} proposed two different architectures: (i) AntSynNET: a pattern-based model and (ii) Combined AntSynNET, i.e., combining the pattern-based model with the distributional embeddings. The Combined AntSynNET comprises two models namely: (1) \emph{AntSynNET + Glove} and (2) \emph{AntSynNET + dLCE}, acquired by using Glove/dLCE embeddings. We use the scores reported in the published paper.

\begin{table*}[htbp]
    \centering
    \resizebox{0.78\linewidth}{!}{%
        \begin{tabular}{l|l|c| c |c c c |c c c |c c c}
            \hline
            
            \multicolumn{2}{l}{\multirow{8}{*}{}} &\multirow{2}{*}{\textbf{Embeddings}} & \multirow{2}{*}{\textbf{Model}} & \multicolumn{3}{c|}{Adjective} & \multicolumn{3}{c|}{Verb} & \multicolumn{3}{c}{Noun} \\
            \cline{5-13}
            \multicolumn{2}{l}{} & \multirow{3}{*}{} & & P & R & F1 & P & R & F1 & P & R & F1 \\
            \hline
            \hline
            \multicolumn{2}{l}{} & \multirow{3}{*}{} & Discourse Markers & 0.717 & 0.717 & 0.717 & 0.789 & 0.787 & 0.788 & 0.833 & 0.831 & 0.832 \\
            \multicolumn{2}{l}{} &  \multirow{1}{*}{None}& Symmetric Patterns & 0.730 & 0.706 & 0.718 & 0.560 & 0.609 & 0.584 & 0.625 & 0.393 & 0.482 \\
            \multicolumn{2}{l}{} &  & AntSynNET & 0.764 & 0.788 & 0.776 & 0.741 & 0.833 & 0.784 & 0.804 & 0.851 & 0.827 \\
            \hline
            \hline
            \multicolumn{2}{l}{} &  & \Direct{} Baseline & 0.700 & 0.619 & 0.657 & 0.634 & 0.630 & 0.632 & 0.682 & 0.647 & 0.664 \\
            \multicolumn{2}{l}{} & \multirow{1}{*}{Glove} & AntSynNET & 0.750 & 0.798 & 0.773 & 0.717 & 0.826 & 0.768 & 0.807 & 0.827 & 0.817 \\
            \multicolumn{2}{l}{} &  & Our-Model & \textbf{0.854} & \textbf{0.917} & \textbf{0.884} & \textbf{0.871} & \textbf{0.912} & \textbf{0.891} & \textbf{0.823} & \textbf{0.866} & \textbf{0.844} \\
            \hline
            \hline
            \multicolumn{2}{l}{} &  & \Direct{} Baseline & 0.897 & 0.897 & 0.897 & 0.857 & 0.833 & 0.845 & 0.890 & 0.859 & 0.874 \\
            \multicolumn{2}{l}{} & \multirow{1}{*}{dLCE} & AntSynNET & 0.763 & 0.807 & 0.784 & 0.743 & 0.815 & 0.777 & 0.816 & 0.898 & 0.855 \\
            \multicolumn{2}{l}{} &  & Our-Model & \textbf{0.912} & \textbf{0.944} & \textbf{0.928} & \textbf{0.899} & \textbf{0.944} & \textbf{0.921} & \textbf{0.905} & \textbf{0.918} & \textbf{0.911}\\
            \hline
            
        \end{tabular}%
    }
    \caption{Antonym/Synonym distinction performance comparison against baseline models}
    \label{tab:Results}
\end{table*}

\subsection{Main Results}

The results for the proposed model are shown in Table~\ref{tab:Results}. Here, we use two different sets of pre-trained embeddings for \Distiller{}, i.e., Glove and dLCE. The results are accordingly compared with the corresponding baseline models using the same settings. 

Comparing the results for the Glove embeddings, our model outperforms the \Direct{} baseline and previously best performing models for all three classes by a significant margin. We observe that for nouns the improvement in performance is relatively lower as compared with the verbs and adjectives.
It is due to the effect of polysemy, which is more dominant among nouns. The unsupervised embeddings are unable to handle polysemous words and mostly the embedding vectors are oriented in the direction of most common sense (as explained in detail in the section~\ref{error:analysis}). This finding is also aligned with that of~\cite{DBLP:conf/ijcnlp/ScheibleWS13}, which states that synonym and antonym pairs conforming to verbs and adjectives have relatively high contextual clues compared with that of nouns. Overall results for our model with Glove embeddings show that even getting started with entirely unsupervised embeddings having a mixture of different lexical-semantic relations, the \Distiller{} is able to distill the task-specific information and outperform the previous state-of-the-art models by a large margin.

Especially noteworthy is the performance of our model with \Distiller{} 
trained on dLCE. It simply outperforms the best performing model trained 
using dLCE, yielding a much higher value of F1 
across all three word classes. Compared with the previous state-of-the-art, it improves the 
F1 score by 18\%, 17\% and 6\% for adjectives, verbs and nouns respectively. 
This drastic improvement in performance explains that in contrary to the Glove embeddings that contains a blend of lexical-semantic 
information, the dLCE embeddings are enriched with more distinguishing 
information and the \Distiller{} helps to effectively distill the most relevant task-specific information for antonym/synonym distinction. 

Such promising results strengthen our claim that the pre-trained embeddings contain a blend of lexico-semantic information and we can acquire task-specific information by projecting them to low-dimensional subspaces.

\subsection{Detailed Performance of Our Model}

Table~\ref{tab:Feature_Analysis} shows the result of our model with different set of features and pre-trained embeddings. We also report the performance of our model with random vectors in order to analyze the performance gained by the unsupervised contextual embeddings.

While the \Distiller{} contributes the most distinction power to our model, we also analyze the results of applying either the \emph{distributional} feature or the \emph{prefix} feature, as well as both of them. It can be seen that the \emph{prefix} feature slightly improved the performance for nouns, yielding a higher F1 score, whereas, for adjectives and verbs, the increase in performance was not so significant. It confirms that the distilled embeddings have a strong distinction power and even the best known features for antonymy detection, i.e., \emph{prefix}, had a very little impact on overall performance. The \emph{distributional} feature slightly improved the performance of the nouns, however, it deteriorated the performance of the verbs and the adjectives. For most cases, using all the features at the same time (i.e., \Distiller{} + \emph{distributional} + \emph{prefix}) had a complementary effect on the model performance, it helped the model to reinforce the decision for confident candidate pairs.

Note that the \Distiller{} trained using random vectors results in a significant reduction in performance (first row in Table~\ref{tab:Feature_Analysis}). It shows that the \Distiller{} is not governed by input bias, rather it distills information contained in unsupervised contextual embeddings. Overall results in Table~\ref{tab:Feature_Analysis} show that our model outperforms the \Direct{} baseline and the previous state-of-the-art model (AntSynNET) for all sets of pre-trained embeddings. These results, moreover, confirm that the dLCE embeddings are indeed customized for antonym/synonym distinction task and on contrary to the AntSynNET, the proposed framework along with the \Distiller{} is able to use this information more effectively.

\subsubsection{Recognizing Irrelevant Pairs}

Although, our current problem setting is not explicitly designed for multi-class classification, our model has an implicit ability to recognize irrelevant pairs. For irrelevant pairs, the resultant projections of words in SYN and ANT sub-spaces end up in a narrow region far from their synonyms and antonyms. For analysis, we augmented the dataset by adding an equal proportion of randomly selected words as “irrelevant” pairs and retrained the classifier in phase II for multi-class classification. With Glove embeddings, our model can distinguish three classes (antonyms, synonyms and irrelevant pairs) with F1 = 0.813, 0.775, and 0.818 for adjectives, verbs, and nouns respectively. These scores are better than the binary classification in previous state-of-the-art (i.e. F1= 0.773, 0.768 and 0.817 for AntSynNET), shown in Table \ref{tab:Results}.

\subsection{Training Time}
We analyzed the pre-processing and running time of our model in comparison with the previous best performing pattern-based approach, i.e., AntSynNET. 

Our model doesn't require any pre-processing, whereas, on our machine (explained
in section~\ref{experimental_settings}) the AntSynNET takes more than 1 week to
parse each sentence in the wiki dump. Moreover, our model is more than 100 times
faster to train (e.g., AntSynNET takes more than 50 hours for model training,
whereas, our model takes less than half an hour).

\begin{table*}[htbp]
    \centering
    \resizebox{.85\linewidth}{!}{
        \begin{tabular}{c|l|ccc|ccc|ccc}
           \hline
             \multirow{2}{*}{\textbf{Embeddings}} &  \multirow{2}{*}{\textbf{Model}} &  & Adjective &  &  & Verb &  &  & Noun &  \\ 
            \cline{3-11}
            \multicolumn{1}{c|}{} & \multicolumn{1}{l|}{} & \multicolumn{1}{c}{P} & \multicolumn{1}{c}{R} & \multicolumn{1}{c|}{F1} & \multicolumn{1}{c}{P} & \multicolumn{1}{c}{R} & \multicolumn{1}{c|}{F1} & \multicolumn{1}{c}{P} & \multicolumn{1}{c}{R} & \multicolumn{1}{c}{F1} \\ \hline \hline

            \multicolumn{1}{c|}{Random Vectors} & \multicolumn{1}{l|}{Distiller + \emph{distributional} + \emph{prefix}} & \multicolumn{1}{c}{0.639} & \multicolumn{1}{c}{0.769} & \multicolumn{1}{c|}{0.698} & \multicolumn{1}{c}{0.719} & \multicolumn{1}{c}{0.833} & \multicolumn{1}{c|}{0.771} & \multicolumn{1}{c}{0.672} & \multicolumn{1}{c}{0.736} & \multicolumn{1}{c}{0.702} \\ \hline \hline

            \multicolumn{1}{c|}{} & \multicolumn{1}{l|}{Direct Baseline} & \multicolumn{1}{c}{0.700} & \multicolumn{1}{c}{0.619} & \multicolumn{1}{c|}{0.657} & \multicolumn{1}{c}{0.634} & \multicolumn{1}{c}{0.630} & \multicolumn{1}{c|}{0.632} & \multicolumn{1}{c}{0.682} & \multicolumn{1}{c}{0.647} & \multicolumn{1}{c}{0.664} \\
            \multicolumn{1}{c|}{} & \multicolumn{1}{l|}{AntSynNET} & \multicolumn{1}{c}{0.750} & \multicolumn{1}{c}{0.798} & \multicolumn{1}{c|}{0.773} & \multicolumn{1}{c}{0.717} & \multicolumn{1}{c}{0.826} & \multicolumn{1}{c|}{0.768} & \multicolumn{1}{c}{0.807} & \multicolumn{1}{c}{0.827} & \multicolumn{1}{c}{0.817} \\
            \multicolumn{1}{c|}{} & \multicolumn{1}{l|}{Distiller} & \multicolumn{1}{c}{\textbf{0.859}} & \multicolumn{1}{c}{0.912} & \multicolumn{1}{c|}{\textbf{0.885}} & \multicolumn{1}{c}{0.866} & \multicolumn{1}{c}{\textbf{0.914}} & \multicolumn{1}{c|}{0.889} & \multicolumn{1}{c}{\textbf{0.823}} & \multicolumn{1}{c}{0.848} & \multicolumn{1}{c}{0.835} \\
            \multicolumn{1}{c|}{Glove} & \multicolumn{1}{l|}{Distiller + \emph{distributional}} & \multicolumn{1}{c}{0.852} & \multicolumn{1}{c}{0.914} & \multicolumn{1}{c|}{0.884} & \multicolumn{1}{c}{\textbf{0.873}} & \multicolumn{1}{c}{0.910} & \multicolumn{1}{c|}{\textbf{0.891}} & \multicolumn{1}{c}{0.821} & \multicolumn{1}{c}{0.853} & \multicolumn{1}{c}{0.837} \\
            \multicolumn{1}{c|}{} & \multicolumn{1}{l|}{Distiller + \emph{prefix}} & \multicolumn{1}{c}{0.855} & \multicolumn{1}{c}{0.916} & \multicolumn{1}{c|}{0.884} & \multicolumn{1}{c}{0.862} & \multicolumn{1}{c}{0.913} & \multicolumn{1}{c|}{0.887} & \multicolumn{1}{c}{0.822} & \multicolumn{1}{c}{\textbf{0.868}} & \multicolumn{1}{c}{\textbf{0.844}} \\
            \multicolumn{1}{c|}{} & \multicolumn{1}{l|}{Distiller + \emph{distributional} + \emph{prefix}} & \multicolumn{1}{c}{0.854} & \multicolumn{1}{c}{\textbf{0.917}} & \multicolumn{1}{c|}{0.884} & \multicolumn{1}{c}{0.871} & \multicolumn{1}{c}{0.912} & \multicolumn{1}{c|}{\textbf{0.891}} & \multicolumn{1}{c}{\textbf{0.823}} & \multicolumn{1}{c}{0.866} & \multicolumn{1}{c}{\textbf{0.844}} \\ \hline\hline

            \multicolumn{1}{c|}{} & \multicolumn{1}{l|}{Direct Baseline} & \multicolumn{1}{c}{0.897} & \multicolumn{1}{c}{0.897} & \multicolumn{1}{c|}{0.897} & \multicolumn{1}{c}{0.857} & \multicolumn{1}{c}{0.833} & \multicolumn{1}{c|}{0.845} & \multicolumn{1}{c}{0.890} & \multicolumn{1}{c}{0.859} & \multicolumn{1}{c}{0.874} \\
            \multicolumn{1}{c|}{} & \multicolumn{1}{l|}{AntSynNET} & \multicolumn{1}{c}{0.763} & \multicolumn{1}{c}{0.807} & \multicolumn{1}{c|}{0.784} & \multicolumn{1}{c}{0.743} & \multicolumn{1}{c}{0.815} & \multicolumn{1}{c|}{0.777} & \multicolumn{1}{c}{0.816} & \multicolumn{1}{c}{0.898} & \multicolumn{1}{c}{0.855} \\
            \multicolumn{1}{c|}{} & \multicolumn{1}{l|}{Distiller} & \multicolumn{1}{c}{\textbf{0.920}} & \multicolumn{1}{c}{0.938} & \multicolumn{1}{c|}{0.929} & \multicolumn{1}{c}{\textbf{0.904}} & \multicolumn{1}{c}{0.930} & \multicolumn{1}{c|}{0.917} & \multicolumn{1}{c}{0.902} & \multicolumn{1}{c}{0.909} & \multicolumn{1}{c}{0.906} \\
            \multicolumn{1}{c|}{dLCE} & \multicolumn{1}{l|}{Distiller + \emph{distributional}} & \multicolumn{1}{c}{0.905} & \multicolumn{1}{c}{0.945} & \multicolumn{1}{c|}{0.924} & \multicolumn{1}{c}{0.886} & \multicolumn{1}{c}{0.938} & \multicolumn{1}{c|}{0.911} & \multicolumn{1}{c}{0.893} & \multicolumn{1}{c}{0.918} & \multicolumn{1}{c}{0.905} \\
            \multicolumn{1}{l|}{} & \multicolumn{1}{l|}{Distiller + \emph{prefix}} & \multicolumn{1}{l}{0.914} & \multicolumn{1}{c}{\textbf{0.947}} & \multicolumn{1}{l|}{\textbf{0.930}} & \multicolumn{1}{l}{0.893} & \multicolumn{1}{l}{\textbf{0.944}} & \multicolumn{1}{l|}{0.918} & \multicolumn{1}{l}{0.894} & \multicolumn{1}{l}{\textbf{0.925}} & \multicolumn{1}{l}{0.909} \\
            \multicolumn{1}{l|}{} & \multicolumn{1}{l|}{Distiller + \emph{distributional} + \emph{prefix}} & \multicolumn{1}{c}{0.912} & \multicolumn{1}{c}{0.944} & \multicolumn{1}{l|}{0.928} & \multicolumn{1}{c}{0.899} & \multicolumn{1}{c}{\textbf{0.944}} & \multicolumn{1}{c|}{\textbf{0.921}} & \multicolumn{1}{c}{\textbf{0.905}} & \multicolumn{1}{c}{0.918} & \multicolumn{1}{c}{\textbf{0.911}} \\ \hline\hline

            \multicolumn{1}{l|}{} & \multicolumn{1}{l|}{Direct Baseline} & \multicolumn{1}{c}{0.676} & \multicolumn{1}{c}{0.589} & \multicolumn{1}{c|}{0.63} & \multicolumn{1}{c}{0.661} & \multicolumn{1}{c}{0.665} & \multicolumn{1}{c|}{0.663} & \multicolumn{1}{c}{0.673} & \multicolumn{1}{c}{0.633} & \multicolumn{1}{c}{0.653} \\
            \multicolumn{1}{l|}{} & \multicolumn{1}{l|}{Distiller} & \multicolumn{1}{c}{\textbf{0.839}} & \multicolumn{1}{c}{0.896} & \multicolumn{1}{c|}{0.866} & \multicolumn{1}{c}{\textbf{0.871}} & \multicolumn{1}{c}{0.926} & \multicolumn{1}{c|}{0.898} & \multicolumn{1}{c}{\textbf{0.833}} & \multicolumn{1}{c}{0.869} & \multicolumn{1}{c}{0.850} \\
            \multicolumn{1}{c|}{ELMO} & \multicolumn{1}{l|}{Distiller + \emph{distributional}} & \multicolumn{1}{c}{0.835} & \multicolumn{1}{c}{0.898} & \multicolumn{1}{c|}{0.866} & \multicolumn{1}{c}{0.870} & \multicolumn{1}{c}{0.929} & \multicolumn{1}{c|}{0.899} & \multicolumn{1}{c}{0.831} & \multicolumn{1}{c}{0.875} & \multicolumn{1}{c}{0.853} \\
            \multicolumn{1}{l|}{} & \multicolumn{1}{l|}{Distiller + \emph{prefix}} & \multicolumn{1}{c}{0.835} & \multicolumn{1}{c}{\textbf{0.909}} & \multicolumn{1}{c|}{\textbf{0.871}} & \multicolumn{1}{c}{0.868} & \multicolumn{1}{c}{0.930} & \multicolumn{1}{c|}{0.898} & \multicolumn{1}{c}{0.831} & \multicolumn{1}{c}{0.876} & \multicolumn{1}{c}{0.852} \\
            \multicolumn{1}{l|}{} & Distiller + \emph{distributional} + \emph{prefix} & \multicolumn{1}{c}{\textbf{0.839}} & \multicolumn{1}{c}{0.905} & \multicolumn{1}{c|}{\textbf{0.871}} & \multicolumn{1}{c}{0.869} & \multicolumn{1}{c}{\textbf{0.933}} & \multicolumn{1}{c|}{\textbf{0.900}} & \multicolumn{1}{c}{0.832} & \multicolumn{1}{c}{\textbf{0.884}} & \multicolumn{1}{c}{\textbf{0.857}} \\ \hline
    \end{tabular}}
    \caption{Performance of proposed model under different settings}
    \label{tab:Feature_Analysis}
\end{table*}

\subsection{Language-Agnostic Model} 
We verify the language agnostic property of our model by testing its performance for antonym/synonym distinction task on another language. We manually curated a dataset for the Urdu language using linguistic 
resource and expert verification. It consists of 750 instances with an equal proportion of antonyms and synonyms. We split the dataset into 70\% train, 25\% test and 5\% validation set. We use fasttext for Urdu language~\cite{DBLP:journals/corr/JoulinGBDJM16} as the pre-trained embeddings. The performance of our model for the Urdu language 
is shown in Table~\ref{tab:result_urdu}. Huge improvement in the performance
compared with the \Direct{} baseline shows that the proposed model is
language-agnostic and has potential to work for other languages provided with
the availability of pre-trained embeddings and a minimal set of training seeds.  

\begin{table}[htbp]
    \centering
    \resizebox{.70\linewidth}{!}{
      \begin{tabular}{c| c c c}
        \hline
            \textbf{Model} & \textbf{P} & \textbf{R} & \textbf{F1}\\
            \hline
            \Direct{} baseline            & 0.623          & 0.533          & 0.575\\
            Our Model & \textbf{0.897} & \textbf{0.867} & \textbf{0.881} \\
            \hline
    \end{tabular}}
    \caption{Antonym/Synonym distinction for Urdu language}
    \label{tab:result_urdu}
\end{table}

\subsection{Analyses of Errors and Distilled Embeddings}
In this section, we perform a detailed error analysis, followed by analyzing the
probable antonyms/synonyms captured via distilled embeddings. We also list the
key advantages of the \Distiller{}.

\subsubsection{Error Cases}
\label{error:analysis}
We randomly selected 50 error cases from our model with Glove embeddings for
analysis. We categorized these errors into three distinct categories: (i) 76\% errors are caused by the polysemous words (ii) 8\% errors are caused by the rarely used words (iii) 16\% other errors.

A major portion of these errors are caused by the inherent limitation of the unsupervised embeddings, i.e., its inability to deal with multiple senses and rare tokens. For case (i), we analysed synonym pairs, e.g., \texttt{(author, generator)}, and antonym pairs, e.g., \texttt{(frog, english)}. Here, at least one word is polysemous with embeddings oriented in direction of common sense, which is different from the sense required in antonym/synonym relation. This phenomenon is more dominant for the nouns. For example, \texttt{generator}'s vector is close to electrical device; \texttt{frog} is commonly used for reptile, however, in the given pair it represents a French person. For case (ii), the embeddings corresponding to the rarely used tokens are not adequately trained, e.g., \texttt{natal} is a rarely used word for \texttt{native}. This results the relation pair to have very low embeddings’ similarity (orthogonal vectors), which limits the \Distiller{} to capture meaningful projections to low-dimensional sub-spaces.

We also observe that increase in word frequency has a deteriorating effect on model performance. This is because most of the high frequency words are polysemous. For analysis, we evenly partitioned the test data (word pairs $(a,b)$) based on the minimum frequency of two words in a large-scale text corpus. For the least frequent 10\% adjective word pairs, the F1 is 0.914, and for the most frequent 10\% word pairs, the F1 drops down to 0.820. A similar trend is observed for nouns and verbs.

\subsubsection{Effectiveness of \Distiller{}}
Pre-trained word embeddings yield a blend of lexical-semantic relations as nearest neighbor of an input word. On contrary, the \Distiller{} provides us the provision to separately analyze just the antonyms and/or synonyms of a given word. We analyzed the nearest neighbors of \texttt{large} using Glove and the distilled embeddings. Results corresponding to Glove embeddings (first column in Table \ref{tab:result_NN}) show that it makes no distinction among lexico-semantic relations, e.g., the antonym \texttt{small} is ranked highest in the list. The results corresponding to the distilled embeddings (column-2) show that the top ranked terms (e.g., \texttt{immense, gigantic}) are indeed the synonyms. Likewise, the results in column-3 show that the top ranked terms by the
\Distiller{} (e.g., \texttt{slender, thin}) are indeed the antonyms. 
 
\begin{table}[htbp]
    \centering
    \resizebox{.9\linewidth}{!}{
        \begin{tabular}{c | c | c}
            \textbf{Glove} & \textbf{\Distiller{}(Synonyms)} & \textbf{\Distiller{}(Antonyms)}\\
            \hline
            small (antonym)   &   immense     &   slender \\
            larger (synonym)  &   gigantic    &   thin    \\
            smaller (antonym) &   influential &   lightweight \\
            huge (synonym)    &   full        &   small   \\
            sized (other)   &   enormous    &   inconspicuous   \\
            \hline
    \end{tabular}}
    \caption{Top-5 nearest neighbors of the word: \texttt{large} using Glove embeddings vs synonyms and antonyms captured by the distilled embeddings}
    \label{tab:result_NN}
\end{table}

\subsubsection{Key Advantages of \Distiller{}}
The key advantages of \Distiller{} compared with the previous state-of-the-art 
pattern based methods are listed below:
\begin{itemize}
    \item \Distiller{} is not constrained by naive assumptions. For example, the pattern based methods work only if a dependency path exists between candidate antonym/synonym pair, and it is impossible for every feasible synonym and/or antonym pair to co-occur within a sentence. Likewise, the \Distiller{} is not limited by the challenges posed by the overlapping nature of lexico-syntactic patterns, which hinder the performance of pattern based methods.
    
    \item \Distiller{} allows explicitly constraining the sub-spaces. This formulation is best suited for lexico-semantic analysis, as it allows appropriate re-ordering of the sub-spaces to capture relation-specific characteristics.
    
  \item Thanks to the wide availability of pre-trained embeddings with various
    improvements and for most languages, \Distiller{} is a flexible and
    efficient choice for lexical-semantic analysis requiring no
    pre-processing and little linguistic knowledge. 
\end{itemize}


\section{Conclusions and Future Work}
\label{conclusions_}
In this paper, we proposed a novel framework, \Distiller{}, for antonym and
synonym distinction. It employs carefully crafted loss functions to
project the pre-trained embeddings to low-dimensional task-specific sub-spaces
in a performance enhanced fashion. Results show that the \Distiller{}
outperforms previous methods on the antonym and synonym distinction task. In the
future, we
will extend the proposed framework to other lexical-semantic relations (e.g.,
hypernymy detection) and to other embedding models such as the hyperbolic
embeddings~\cite{DBLP:conf/nips/NickelK17}.

\paragraph{Acknowledgments.}
This research was partially funded by ARC DPs 170103710 and 180103411,
and D2DCRC DC25002 and DC25003, and NSFC Grant 61872446.

\bibliography{aaai2019}

\begin{thebibliography}{}

\bibitem[\protect\citeauthoryear{Adel and
  Sch{\"{u}}tze}{2014}]{DBLP:conf/emnlp/AdelS14}
Adel, H., and Sch{\"{u}}tze, H.
\newblock 2014.
\newblock Using mined coreference chains as a resource for a semantic task.
\newblock In {\em {EMNLP}},  1447--1452.
\newblock {ACL}.

\bibitem[\protect\citeauthoryear{Bordes \bgroup et al\mbox.\egroup
  }{2013}]{DBLP:conf/nips/BordesUGWY13}
Bordes, A.; Usunier, N.; Garc{\'{\i}}a{-}Dur{\'{a}}n, A.; Weston, J.; and
  Yakhnenko, O.
\newblock 2013.
\newblock Translating embeddings for modeling multi-relational data.
\newblock In {\em {NIPS}},  2787--2795.

\bibitem[\protect\citeauthoryear{Chen and
  Guestrin}{2016}]{DBLP:conf/kdd/ChenG16}
Chen, T., and Guestrin, C.
\newblock 2016.
\newblock Xgboost: {A} scalable tree boosting system.
\newblock In {\em {KDD}},  785--794.
\newblock {ACM}.

\bibitem[\protect\citeauthoryear{Fellbaum}{1998}]{fellbaum1998wordnet}
Fellbaum, C.
\newblock 1998.
\newblock {\em WordNet}.
\newblock Wiley Online Library.

\bibitem[\protect\citeauthoryear{Glavas and
  Vulic}{2018}]{DBLP:conf/naacl/GlavasV18}
Glavas, G., and Vulic, I.
\newblock 2018.
\newblock Discriminating between lexico-semantic relations with the
  specialization tensor model.
\newblock In {\em {NAACL-HLT} {(2)}},  181--187.
\newblock Association for Computational Linguistics.

\bibitem[\protect\citeauthoryear{Joulin \bgroup et al\mbox.\egroup
  }{2016}]{DBLP:journals/corr/JoulinGBDJM16}
Joulin, A.; Grave, E.; Bojanowski, P.; Douze, M.; J{\'{e}}gou, H.; and Mikolov,
  T.
\newblock 2016.
\newblock Fasttext.zip: Compressing text classification models.
\newblock {\em CoRR} abs/1612.03651.

\bibitem[\protect\citeauthoryear{Kingma and
  Ba}{2014}]{DBLP:journals/corr/KingmaB14}
Kingma, D.~P., and Ba, J.
\newblock 2014.
\newblock Adam: {A} method for stochastic optimization.
\newblock {\em CoRR} abs/1412.6980.

\bibitem[\protect\citeauthoryear{Lampinen and
  McClelland}{2017}]{DBLP:journals/corr/abs-1710-10280}
Lampinen, A.~K., and McClelland, J.~L.
\newblock 2017.
\newblock One-shot and few-shot learning of word embeddings.
\newblock {\em CoRR} abs/1710.10280.

\bibitem[\protect\citeauthoryear{Lin \bgroup et al\mbox.\egroup
  }{2003}]{DBLP:conf/ijcai/LinZQZ03}
Lin, D.; Zhao, S.; Qin, L.; and Zhou, M.
\newblock 2003.
\newblock Identifying synonyms among distributionally similar words.
\newblock In {\em {IJCAI}},  1492--1493.
\newblock Morgan Kaufmann.

\bibitem[\protect\citeauthoryear{Mikolov \bgroup et al\mbox.\egroup
  }{2013a}]{DBLP:journals/corr/abs-1301-3781}
Mikolov, T.; Chen, K.; Corrado, G.; and Dean, J.
\newblock 2013a.
\newblock Efficient estimation of word representations in vector space.
\newblock {\em CoRR} abs/1301.3781.

\bibitem[\protect\citeauthoryear{Mikolov \bgroup et al\mbox.\egroup
  }{2013b}]{DBLP:conf/nips/MikolovSCCD13}
Mikolov, T.; Sutskever, I.; Chen, K.; Corrado, G.~S.; and Dean, J.
\newblock 2013b.
\newblock Distributed representations of words and phrases and their
  compositionality.
\newblock In {\em {NIPS}},  3111--3119.

\bibitem[\protect\citeauthoryear{Mohammad \bgroup et al\mbox.\egroup
  }{2013}]{DBLP:journals/coling/MohammadDHT13}
Mohammad, S.; Dorr, B.~J.; Hirst, G.; and Turney, P.~D.
\newblock 2013.
\newblock Computing lexical contrast.
\newblock {\em Computational Linguistics} 39(3):555--590.

\bibitem[\protect\citeauthoryear{Nguyen, {Schulte im Walde}, and
  Vu}{2016}]{DBLP:conf/acl/NguyenWV16}
Nguyen, K.~A.; {Schulte im Walde}, S.; and Vu, N.~T.
\newblock 2016.
\newblock Integrating distributional lexical contrast into word embeddings for
  antonym-synonym distinction.
\newblock In {\em {ACL} {(2)}}.
\newblock The Association for Computer Linguistics.

\bibitem[\protect\citeauthoryear{Nguyen, {Schulte im Walde}, and
  Vu}{2017}]{DBLP:conf/eacl/NguyenWV17}
Nguyen, K.~A.; {Schulte im Walde}, S.; and Vu, N.~T.
\newblock 2017.
\newblock Distinguishing antonyms and synonyms in a pattern-based neural
  network.
\newblock In {\em {EACL} {(1)}},  76--85.
\newblock Association for Computational Linguistics.

\bibitem[\protect\citeauthoryear{Nickel and
  Kiela}{2017}]{DBLP:conf/nips/NickelK17}
Nickel, M., and Kiela, D.
\newblock 2017.
\newblock Poincar{\'{e}} embeddings for learning hierarchical representations.
\newblock In {\em {NIPS}},  6341--6350.

\bibitem[\protect\citeauthoryear{Ono, Miwa, and
  Sasaki}{2015}]{DBLP:conf/naacl/OnoMS15}
Ono, M.; Miwa, M.; and Sasaki, Y.
\newblock 2015.
\newblock Word embedding-based antonym detection using thesauri and
  distributional information.
\newblock In {\em {HLT-NAACL}},  984--989.
\newblock The Association for Computational Linguistics.

\bibitem[\protect\citeauthoryear{Pennington, Socher, and
  Manning}{2014}]{DBLP:conf/emnlp/PenningtonSM14}
Pennington, J.; Socher, R.; and Manning, C.~D.
\newblock 2014.
\newblock Glove: Global vectors for word representation.
\newblock In {\em {EMNLP}},  1532--1543.
\newblock {ACL}.

\bibitem[\protect\citeauthoryear{Peters \bgroup et al\mbox.\egroup
  }{2018}]{DBLP:conf/naacl/PetersNIGCLZ18}
Peters, M.~E.; Neumann, M.; Iyyer, M.; Gardner, M.; Clark, C.; Lee, K.; and
  Zettlemoyer, L.
\newblock 2018.
\newblock Deep contextualized word representations.
\newblock In {\em {NAACL-HLT}},  2227--2237.
\newblock Association for Computational Linguistics.

\bibitem[\protect\citeauthoryear{Pham, Lazaridou, and
  Baroni}{2015}]{DBLP:conf/acl/PhamLB15}
Pham, N.~T.; Lazaridou, A.; and Baroni, M.
\newblock 2015.
\newblock A multitask objective to inject lexical contrast into distributional
  semantics.
\newblock In {\em {ACL} {(2)}},  21--26.
\newblock The Association for Computer Linguistics.

\bibitem[\protect\citeauthoryear{Rajana \bgroup et al\mbox.\egroup
  }{2017}]{DBLP:conf/starsem/RajanaCAS17}
Rajana, S.; Callison{-}Burch, C.; Apidianaki, M.; and Shwartz, V.
\newblock 2017.
\newblock Learning antonyms with paraphrases and a morphology-aware neural
  network.
\newblock In {\em *SEM},  12--21.
\newblock Association for Computational Linguistics.

\bibitem[\protect\citeauthoryear{Roth and {Schulte im
  Walde}}{2014}]{DBLP:conf/acl/RothW14}
Roth, M., and {Schulte im Walde}, S.
\newblock 2014.
\newblock Combining word patterns and discourse markers for paradigmatic
  relation classification.
\newblock In {\em {ACL} {(2)}},  524--530.
\newblock The Association for Computer Linguistics.

\bibitem[\protect\citeauthoryear{Rothe and
  Sch{\"{u}}tze}{2016}]{DBLP:conf/acl/RotheS16}
Rothe, S., and Sch{\"{u}}tze, H.
\newblock 2016.
\newblock Word embedding calculus in meaningful ultradense subspaces.
\newblock In {\em {ACL} {(2)}}.
\newblock The Association for Computer Linguistics.

\bibitem[\protect\citeauthoryear{Scheible, {Schulte im Walde}, and
  Springorum}{2013}]{DBLP:conf/ijcnlp/ScheibleWS13}
Scheible, S.; {Schulte im Walde}, S.; and Springorum, S.
\newblock 2013.
\newblock Uncovering distributional differences between synonyms and antonyms
  in a word space model.
\newblock In {\em {IJCNLP}},  489--497.
\newblock Asian Federation of Natural Language Processing / {ACL}.

\bibitem[\protect\citeauthoryear{Schwartz, Reichart, and
  Rappoport}{2015}]{DBLP:conf/conll/SchwartzRR15}
Schwartz, R.; Reichart, R.; and Rappoport, A.
\newblock 2015.
\newblock Symmetric pattern based word embeddings for improved word similarity
  prediction.
\newblock In {\em CoNLL},  258--267.
\newblock {ACL}.

\bibitem[\protect\citeauthoryear{Turney and
  Pantel}{2010}]{DBLP:journals/jair/TurneyP10}
Turney, P.~D., and Pantel, P.
\newblock 2010.
\newblock From frequency to meaning: Vector space models of semantics.
\newblock {\em J. Artif. Intell. Res.} 37:141--188.

\bibitem[\protect\citeauthoryear{Vulic and
  Mrksic}{2018}]{DBLP:conf/naacl/VulicM18}
Vulic, I., and Mrksic, N.
\newblock 2018.
\newblock Specialising word vectors for lexical entailment.
\newblock In {\em {NAACL-HLT}},  1134--1145.
\newblock Association for Computational Linguistics.

\bibitem[\protect\citeauthoryear{Vulic}{2018}]{DBLP:conf/rep4nlp/Vulic18}
Vulic, I.
\newblock 2018.
\newblock Injecting lexical contrast into word vectors by guiding vector space
  specialisation.
\newblock In {\em Rep4NLP@ACL},  137--143.
\newblock Association for Computational Linguistics.

\bibitem[\protect\citeauthoryear{Yoon \bgroup et al\mbox.\egroup
  }{2016}]{DBLP:conf/naacl/YoonSPP16}
Yoon, H.; Song, H.; Park, S.; and Park, S.
\newblock 2016.
\newblock A translation-based knowledge graph embedding preserving logical
  property of relations.
\newblock In {\em {HLT-NAACL}},  907--916.
\newblock The Association for Computational Linguistics.

\end{thebibliography}
\bibliographystyle{aaai} 
\end{document}